\documentclass[letterpaper]{article} 
\usepackage{aaai24}  
\usepackage{times}  
\usepackage{helvet}  
\usepackage{courier}  
\usepackage[hyphens]{url}  
\usepackage{graphicx} 
\usepackage{natbib}  
\usepackage{caption} 
\usepackage{amsmath}
\usepackage{subfigure}
\usepackage{algorithm}
\usepackage{algorithmic}

\usepackage{newfloat}
\usepackage{listings}
\DeclareCaptionStyle{ruled}{labelfont=normalfont,labelsep=colon,strut=off} 
\floatstyle{ruled}
\newfloat{listing}{tb}{lst}{}
\floatname{listing}{Listing}
\title{BEVGPT: Generative Pre-trained Large Model for Autonomous Driving Prediction, Decision-Making, and Planning}
\author{
    Pengqin Wang\textsuperscript{\rm 1, \rm 2},
    Meixin Zhu\textsuperscript{\rm 1, \rm 2, \rm 5}\thanks{ Corresponding author},
    Hongliang Lu\textsuperscript{\rm 2},
    Hui Zhong\textsuperscript{\rm 2},
    Xianda Chen\textsuperscript{\rm 2},\\
    Shaojie Shen\textsuperscript{\rm 1},
    Xuesong Wang\textsuperscript{\rm 3},
    Yinhai Wang\textsuperscript{\rm 4}
}
\affiliations{
    \textsuperscript{\rm 1} The Hong Kong University of Science and Technology\\
    \textsuperscript{\rm 2} The Hong Kong University of Science and Technology (Guangzhou)\\
    \textsuperscript{\rm 3} Tongji University\\
    \textsuperscript{\rm 4} University of Washington\\
    \textsuperscript{\rm 5} Guangdong Provincial Key Lab of Integrated Communication, Sensing and Computation for Ubiquitous Internet of Things\\

    pwangas@connect.ust.hk,
    meixin@ust.hk,
    \{hlu592, hzhong638, xchen595\}@connect.hkust-gz.edu.cn,\\
    eeshaojie@ust.hk,
    wangxs@tongji.edu.cn,
    yinhai@uw.edu
}

\usepackage{bibentry}

\begin{document}

\maketitle

\begin{abstract}
Prediction, decision-making, and motion planning are essential for autonomous driving. In most contemporary works, they are considered as individual modules or combined into a multi-task learning paradigm with a shared backbone but separate task heads. However, we argue that they should be integrated into a comprehensive framework. Although several recent approaches follow this scheme, they suffer from complicated input representations and redundant framework designs. More importantly, they can not make long-term predictions about future driving scenarios. To address these issues, we rethink the necessity of each module in an autonomous driving task and incorporate only the required modules into a minimalist autonomous driving framework. We propose BEVGPT, a generative pre-trained large model that integrates driving scenario prediction, decision-making, and motion planning. The model takes the bird's-eye-view (BEV) images as the only input source and makes driving decisions based on surrounding traffic scenarios. To ensure driving trajectory feasibility and smoothness, we develop an optimization-based motion planning method. We instantiate BEVGPT on Lyft Level 5 Dataset and use Woven Planet L5Kit for realistic driving simulation. The effectiveness and robustness of the proposed framework are verified by the fact that it outperforms previous methods in 100\% decision-making metrics and 66\% motion planning metrics. Furthermore, the ability of our framework to accurately generate BEV images over the long term is demonstrated through the task of driving scenario prediction. To the best of our knowledge, this is the first generative pre-trained large model for autonomous driving prediction, decision-making, and motion planning with only BEV images as input.
\end{abstract}

\section{1 Introduction}
Autonomous driving vehicles are intelligent systems that integrate technologies for prediction, decision-making, and planning \cite{wei2014behavioral, ding2021epsilon}. A widely accepted solution is to divide these tasks into different modules and develop task-specific models for each module \cite{zhan2017spatially, 9197302}. In this scheme, the future trajectories of surrounding traffic agents are first predicted based on environmental information. Then the predicted agents' motion as well as map segmentation is used to make driving decisions and plan future trajectories for self-driving vehicles. However, by taking this modular approach, the system can be easily affected by accumulative errors across different modules. An alternative paradigm is multi-task learning with a shared feature extracting backbone but separate task heads \cite{liang2022effective, chen2022learning}. Although this approach reduces model size and complexity and achieves faster computational speed and less computing overheads, it may suffer from negative transfer because sharing information with an unrelated task might actually harm model performance \cite{ruder2017overview}. 

Instead, we argue that the various modules of autonomous driving should be integrated into a comprehensive framework. Recent research demonstrates that the bird's-eye-view (BEV) perspective has tremendous potential for autonomous driving systems \cite{li2023powerbev, hu2023planning}. BEV provides an accurate representation of neighboring traffic conditions for vision-centric perception \cite{li2022bevformer, akan2022stretchbev}. 
However, existing methods for BEV generation require complicated input representations. In addition, these approaches lack the capacity for long-term modeling, making their predictions of future BEV unreliable.

\begin{figure*}[!ht]
  \centering
  \includegraphics[width=0.9\textwidth]{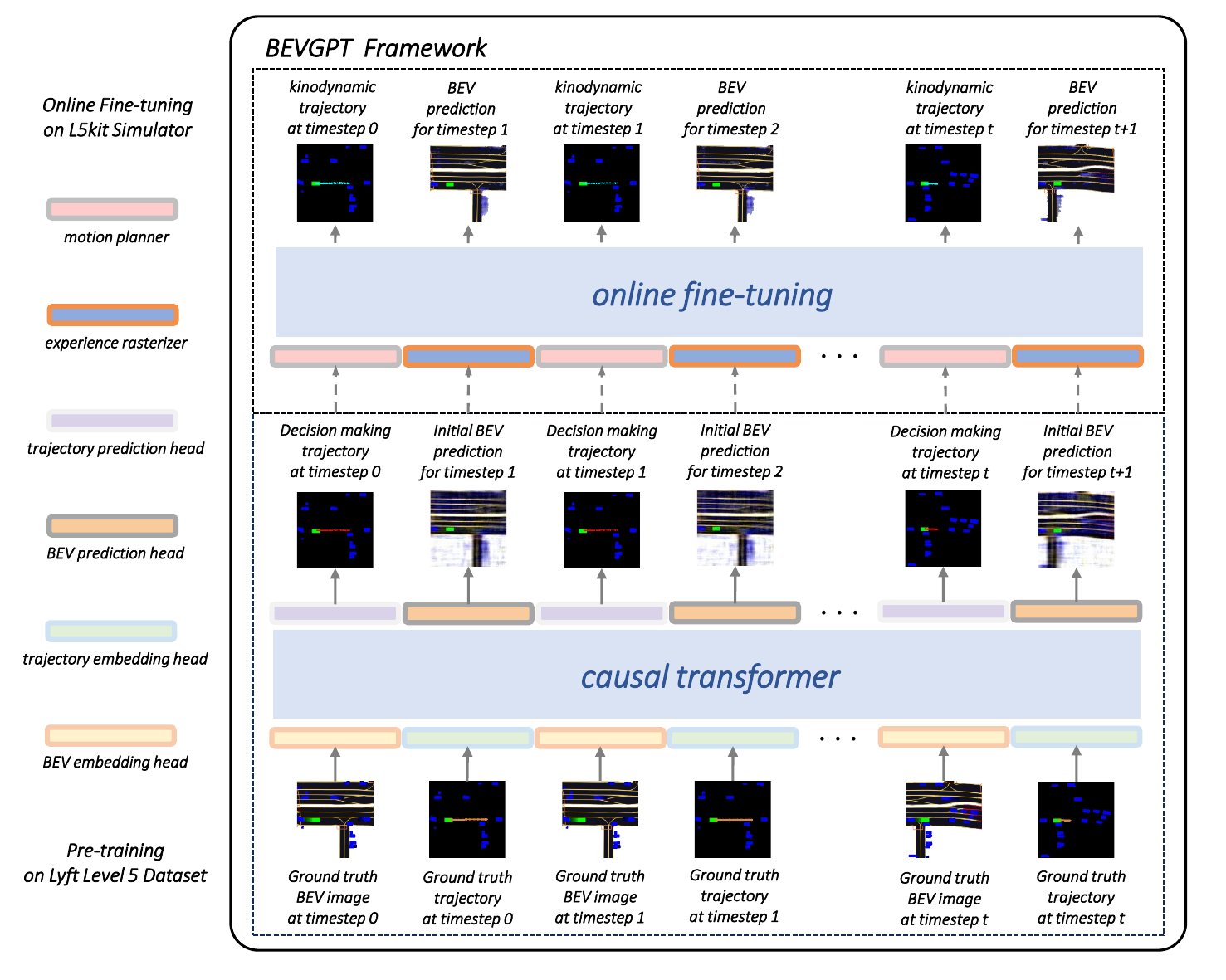}
  \caption{The two-stage training procedure of our proposed BEVGPT framework. In the pre-training stage, we use a large number of self-driving data to train a causal transformer. The objective is to learn driving scenario prediction and decision-making. The model is high-capacity, which can decide the future trajectory for 4 seconds and predict the future driving scenarios for up to 6 seconds. In the online fine-tuning stage, we adapt the trained causal transformer towards kinodynamic motion planning and accurate BEV generation.}\label{fig:BEVGPT}
\end{figure*}

In this paper, we propose a comprehensive framework named BEVGPT that integrates prediction, decision-making, and motion planning in a single Generative Pre-trained Transformer (GPT) with only BEV images as input. As shown in Figure~\ref{fig:BEVGPT}, we employ a two-stage training procedure. First, we use massive self-driving data to train a causal transformer. Subsequently, we fine-tune the model through online learning using a realistic simulator.
Specifically, the objective of the pre-training stage is to learn driving scenario prediction and decision-making, i.e., BEV generation and ego-forecasting in our self-driving task. The model is high-capacity, which can decide future trajectory for 4 seconds and predict future driving scenarios for up to 6 seconds. This is followed by an online fine-tuning stage, in which we adapt the trained causal transformer towards kinodynamic motion planning and accurate BEV prediction. In the fine-tuning stage, a motion planner is designed to generate a smooth and feasible trajectory for the self-driving vehicle. Furthermore, an experience rasterizer is developed to help the model process static information of driving scenarios, such as lanes and intersections.

We instantiate BEVGPT on Lyft Level 5 Dataset \cite{houston2021one}. The Woven Planet L5Kit simulator is used to provide realistic driving simulations for online fine-tuning. We compare our framework with state-of-the-art decision-making and motion planning baselines. The effectiveness and robustness of the proposed framework are verified by performing better than previous methods in most tasks while similarly well in other tasks.
Moreover, the ability of our framework for accurate long-term BEV generation is demonstrated by the driving scenario prediction task.

Our main contributions can be summarized as follows:

\begin{itemize}
  \item [1)] 
  We propose BEVGPT, a comprehensive framework that integrates multiple autonomous driving tasks (driving scenario prediction, decision-making, and motion planning) using a single model. BEVGPT provides a new paradigm for autonomous driving systems where all tasks are incorporated into a single GPT model with only BEV images as input. It demonstrates the enormous potential of the GPT architecture to be applied in complex and dynamic systems.
  \item [2)]
  We instantiate BEVGPT on Lyft Level 5 Dataset and fine-tune it on the realistic L5Kit simulator. Through hundreds of simulation tests under different real-world driving scenarios, we verify the effectiveness and robustness of the proposed framework. Our approach achieves better performance than previous methods in most metrics while performing at the same level in other metrics. Additionally, our model demonstrates a strong ability to forecast future BEV in challenging and complex driving scenarios.
\end{itemize}

In the following sections, we first introduce related work about decision-making and BEV prediction. Then we illustrate the methodology of framework design, differentially flat vehicle model, polynomial trajectory representation, and motion planning in detail. In the next section, we explain network architecture and hyperparameters, the data extraction process, and the two-stage training procedure including pre-training and online fine-tuning. Afterward, we evaluate the performance of the model. The final section is devoted to discussion and conclusion.

\section{2 Related Work}

\subsection{2.1 Decision Making}

As the center of autonomous driving systems, the decision-making module is a crucial component due to the significance of intelligent reaction to a constantly changing environment. The output of the decision module can be low-dimensional motion control instructions, such as the throttle, speed, acceleration, etc., or it can be high-level instructions, such as the motion primitive and future trajectory, which can be adopted by the subsequent planning module. \citep{vallon2017machine} propose an algorithm to make ``Lane-Changing" decisions for autonomous vehicles with relative position and velocity as inputs. \citep{tami2019machine} present an approach based on machine learning to enable autonomous vehicles to make appropriate decisions and fulfill safe maneuvers in ``Cut-In" scenarios. \citep{chen2015deepdriving} propose an end-to-end paradigm where a CNN is trained to map the input images to key perception indicators to make decision in ``Lane-Keeping" scenarios. \citep{8317923} present a method that uses front-facing camera images to train a CNN for obstacle avoidance and evaluate it in real-world driving scenarios. \citep{codevilla2018end} suggest conditioning imitation learning on high-level command input. They train two networks for processing inputs and switching the command actions respectively. \citep{caltagirone2017lidar} propose a learning-based method to generate the high-level driving path based on the multi-sensor inputs including the LiDAR point cloud, GPS-IMU information, and Google driving directions. \citep{wulfmeier2017large} combine inverse reinforcement learning (IRL) with fully convolutional neural networks (FCN) to learn spatial traversability maps for urban driving. \citep{bansal2018chauffeurnet} suggest exposing the learner to the synthesized data by adding perturbations to the experts and training the driving policy via imitation learning (IL). 

In our work, we follow the high-level representation and define the decision-making task as ego-forecasting. The output decision is the future trajectory points of the autonomous vehicle and is adopted by the subsequent motion planner at the fine-tuning stage.

\subsection{2.2 BEV Prediction}

The bird's-eye-view (BEV) is a natural and intuitive representation that can accurately reflect the driving scenarios. According to different characteristics of task requirements and product scenarios, previous BEV prediction methods can be mainly divided into camera-based and LiDAR-based. LiDAR-based methods generally show better performance than camera-based methods due to their ability to provide accurate depth information of objects in 3D space. \citep{zhou2018voxelnet} propose an end-to-end network to unify feature extraction and bounding box prediction. It divides the point cloud into isometric 3D voxels and converts a set of points within each voxel into a uniform feature representation through a VFE layer. \citep{shi2020pv} combine 3D voxel CNN and PointNet-based networks to discover more discriminative point cloud characteristics and predict future BEV. \citep{yin2021center} propose a powerful center-based 3D object detection and tracking framework based on LiDAR point cloud. \citep{deng2021voxel} design a voxel-based framework that uses 3D convolution backbone to extract point cloud features and a 2D BEV region proposal network for object proposals. \citep{wang2021object} model the 3D object detection task as message transmission on a dynamic graph based on BEV representation and propose a set-to-set distillation method for 3D detection. Compared with LiDAR-based BEV prediction and 3D perception, camera-based methods attract a lot of attention recently because they use only camera images to lift 2D features to 3D and then project the lifted features to BEV representation. \citep{philion2020lift} propose an end-to-end architecture named LSS that directly extracts the BEV representation of a scene based on camera images. \citep{huang2021bevdet} adapt LSS and propose a framework BEVDet to perform 3D object detection tasks in BEV. \citep{hu2021fiery} present a probabilistic future prediction model in BEV from monocular cameras. It extracts BEV features from multi-view cameras and uses recurrent networks to make future predictions in BEV representation. \citep{li2023bevdepth} propose a camera-based BEV 3D object detector with a trustworthy depth estimation. \citep{zhang2022beverse} present a unified framework BEVerse for joint perception and prediction with shared feature extraction and parallel multi-task inference.

We employ BEV as the unified input representation in our work, so we do not investigate the process of generating BEV from multi-sensor inputs. Consequently, our research focuses on how to generate accurate predictions of future BEV based on driving decisions and historical BEV images.

\section{3 Methodology}

In this section, we present the methodology framework, differentially flat vehicle model, piece-wise polynomial trajectory representation, and minimum jerk motion planning in detail.

\subsection{3.1 Framework Design}

We re-evaluate the necessity of each module from human drivers' perspectives to design the autonomous driving system framework:
\begin{itemize}
    \item First, human drivers typically have prior knowledge of the environment map. For instance, when utilizing navigation software, the structure of each road and the location of each intersection are readily apparent.
    \item In addition, human drivers do not explicitly track the agents around them and predict their future trajectories. Instead, they are more concerned with whether the positions on the expected ego trajectory will be occupied by other vehicles in the future. Then they make driving decisions accordingly.
\end{itemize}  

Based on the above considerations, we keep the essential modules that comprise the minimalist autonomous driving system, i.e., decision-making, motion planning, and driving scenario prediction, while removing the object tracking and agent motion forecasting modules from the classical autonomous driving framework.

\begin{figure}[!ht]
  \centering
  \includegraphics[width=0.45\textwidth]{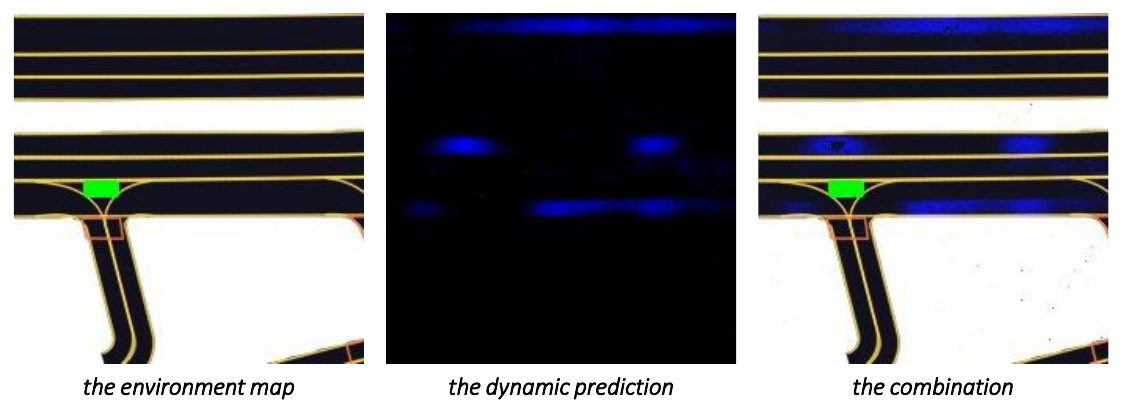}
  \caption{The driving scenario prediction.}\label{fig:combination}
\end{figure}

In our framework, the decision-making outputs are the future positions of the ego vehicle within a horizon of $T=4s$, which is further processed by the motion planner to generate a kinodynamic and smooth trajectory. Considering that static environmental information can be easily obtained from high-definition maps, we pay more attention to the precise prediction of dynamic objects including vehicles and pedestrians. The driving scenario prediction is obtained by the combination of the environment map and the dynamic prediction, as illustrated in Figure ~\ref{fig:combination}.

\subsection{3.2 Vehicle Model}
    \label{section:vehicle_model}

\begin{figure}[!ht]
  \centering
  \includegraphics[width=0.3\textwidth]{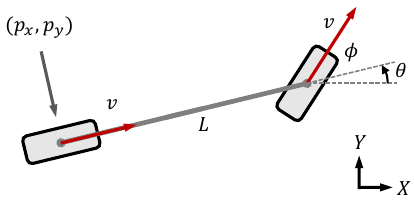}
  \caption{The kinematic bicycle model.}\label{fig:flatness}
\end{figure}

To reasonably represent the motion of the autonomous vehicle and plan a feasible trajectory, we use the kinematic bicycle model, as shown in Figure~\ref{fig:flatness}. The model dynamics can be written as:

\begin{subequations}
	\begin{align}
            & \dot{p_x} = v \cos{\theta}, \\
            & \dot{p_y} = v \sin{\theta}, \\
            & \dot{\theta} = \frac{v}{L} \tan{\phi}, \\
            & \dot{v} = a,
        \end{align}
\end{subequations}
where ${p_x}, {p_y}, {\theta}, {v}$ represent the position at the center of the rear wheels, heading, and speed of the vehicle respectively, while $ {a}$ and ${\phi}$ are the vehicle acceleration and the steering angle of the front wheels. ${L}$ is the wheelbase length of the car. This kinematic bicycle model is shown to be differentially flat when we take 
$ {\sigma} = ({\sigma_1}, {\sigma_2})^{\rm T} = ({p_x}, {p_y})^{\rm T}$
as the flat output \cite{van1998differential, murray1995differential}.
As a result, under the differential flatness characteristic, we can use the flat outputs and their finite derivatives to represent any vehicle states, which contributes to the trajectory representation and motion planning:

\begin{subequations}
	\begin{align}
            & p_x = \sigma_1,  \\
            & p_y = \sigma_2,  \\
            & \theta = \arctan2(\dot{\sigma_2}, \dot{\sigma_1}), \\
		& v = \sqrt{\dot{\sigma_1}^2 + \dot{\sigma_2}^2},  \\
		& a = (  \dot{\sigma_1}\ddot{\sigma_1}+ \dot{\sigma_2}\ddot{\sigma_2}  )   / \sqrt{\dot{\sigma_1}^2 + \dot{\sigma_2}^2} , \label{eq:at}\\
		& \phi = \arctan \left(  (  \dot{\sigma_1}\ddot{\sigma_2}- \dot{\sigma_2}\ddot{\sigma_1}  )L /  ( \dot{\sigma_1}^2 + \dot{\sigma_2}^2 )^{\frac{3}{2}}  \right ).
	\end{align}
\end{subequations}

\subsection{3.3 Trajectory Representation}
    \label{section:trajectory}

We adopt piece-wise polynomial trajectories to represent our differentially flat output, i.e., ${p_x}$ and ${p_y}$. In order to minimize jerk in the later motion planning module, we choose quintic polynomial representation. We assume that the trajectory consists of a total of M segments, each with the same time interval of $\Delta t$. For each dimension, the $n$-th piece polynomial trajectory can be written as:

\begin{equation}
    f_n(t) = \sum_{k=0}^{5} f_{n,k} t^k, t \in [T_{n}, T_{n+1}], 
\end{equation}
for $n = 0, 1, ..., M - 1$ and $T_n = n \Delta t$. And $f_n = [f_{n,0}, f_{n,1}, f_{n,2}, f_{n,3}, f_{n,4}, f_{n,5}]^T$ means the coefficients of the $n$-th polynomial.

\subsection{3.4 Motion Planning}
    \label{section:motion_planning}

The basic requirements of a planning trajectory $f(t)$ include dynamical feasibility and trajectory smoothness. Meanwhile, it is preferable to minimize the control effort. In our case, we choose jerk to represent the control effort. The initial states $\mathbf {p_0}$ and final states $\mathbf {p_M}$ are specified. The decision-making output $p_{n}$ should be contained in the trajectory. In conclusion, the requirements of motion planning give the following minimum jerk problem:

\begin{subequations}
	\begin{align}
            \min_{f(t)} & \label{eq:cost_function}~ \sum_{n=0}^{M-1} J_n = \sum_{n=0}^{M-1}\int_{T_{n}}^{T_{n+1}} {\left \|{f_{n}^{(3)}(t)}\right \|^2} dt, \\
            s.t. 
            ~ & \label{eq:initial_constrain}~f_{0}^{(s)}(0)= \mathbf {p_0}^{(s)}, s \in \{0, 1\},\\
            ~ & \label{eq:final_constrain}~f_{M-1}^{(s)}(T_M)= \mathbf {p_{M}}^{(s)}, s \in \{0, 1\},\\
            ~ & \label{eq:waypoint_constrain}~f_n(T_{n})=p_{n}, \\
            ~ & \label{eq:continue_constrain}~f_{n+1}^{(s)}(T_n)=f_{n}^{(s)}(T_n), s \in \{0, 1, 2\},\\
            ~ & \label{eq:max_constrain}~\left \|f_{n}^{(s)}(T_n)\right \| \leq \mathbf {D_{max}}, s \in \{1, 2\},
        \end{align}
\end{subequations}
where  Eq. \ref{eq:cost_function} is the general form of our motion planning problem, with the goal of minimizing the control effort jerk. Eq. \ref{eq:initial_constrain} and Eq. \ref{eq:final_constrain} denote the initial states $\mathbf {p_0}$ and final states $\mathbf {p_M}$. Eq. \ref{eq:waypoint_constrain} ensures that the initial position for each segment is obtained from the predicted waypoint of the pre-trained causal transformer. Eq. \ref{eq:continue_constrain} provides a guarantee for the continuity between the two segments of the trajectory. Eq. \ref{eq:max_constrain} shows the dynamic constraints where $\mathbf {D_{max}}$ denotes the max speed and the max acceleration. Inspired by \cite{wang2019game}, we use the upper bound of the vehicle acceleration to satisfy the dynamic constraints in practice.

\section{4 Training}

In this section, we first introduce the network architecture and the hyperparameters. Then we illustrate the data extraction procedure. Lastly, we explain the two-stage training process including pre-training and online fine-tuning.

\subsection{4.1 Network and Hyperparameters}

In this study, we employ the GPT architecture \cite{radford2018improving}, which modifies the transformer architecture by adding a causal self-attention mask so that the prediction tokens are auto-regressively generated. Our hyperparameters are shown below in Table \ref{tbl:hyperparameters}.

\begin{table*}[ht]
\renewcommand\arraystretch{1.2}
\caption{Hyperparameters of BEVGPT}
\begin{center}
\begin{small}
\begin{tabular}{ll}\hline
\textbf{Hyperparameter} & \textbf{Value}  \\\hline
BEV dimension    & $224 * 224 * 3$  \\
Waypoint dimension    & $40 * 2$  \\
Number of layers & $6$  \\ 
Number of attention heads    & $16$  \\
Embedding dimension    & $4096$  \\
Nonlinearity function & ReLU \\
Batch size   & $8$ \\
Context length $K$ & $60$ \\
Dropout & $0.1$ \\
Learning rate & $10^{-5}$ \\
Weight decay & $10^{-5}$ \\
Epsilon & $10^{-4}$ \\
Loss weight alpha & 60 degree \\
Learning rate decay & Linear warmup for first $10^4$ training steps \\\hline
\end{tabular}
\end{small}
\label{tbl:hyperparameters}
\end{center}
\end{table*} 

\subsection{4.2 Data Extraction}

Due to the diversity of the self-driving data required by our task, we used the self-driving dataset proposed in \cite{houston2021one}. It comprises more than 1,000 hours of driving data collected by a fleet of 20 autonomous vehicles within a four-month period. We extract vehicle poses, semantic BEV images, and static environment map images from the dataset. Specifically, we first iterate over the dataset and eliminate scenarios shorter than 24 seconds, i.e., 240 frames with the time interval $\Delta t=0.1s$. Then the future target positions of the ego vehicle are extracted as the label for decision-making training. In the end, we record the target positions, current BEV image, next BEV image, and next environment map image for each frame as the training dataset.

\subsection{4.3 Pre-training}

During the pre-training stage, we train BEVGPT for 20 epochs on the extracted dataset. To improve the model's decision-making and prediction capabilities, we design the following loss function using the mean square error (MSE):

\begin{subequations}
	\begin{align}
            \label{eq:decision_loss1}  & loss_{decision} = MSE(p_{de}, p_{gt}),  \\
            \label{eq:prediction_loss1} & loss_{prediction} = 100 \cdot MSE(env_{gt}+bev_{pr}, bev_{gt}),  \\
            \label{eq:loss1} & loss =  \sin{\alpha} \cdot loss_{decision} + \cos{\alpha} \cdot loss_{prediction}, 
	\end{align}
\end{subequations}
where $p_{de}$ and $p_{gt}$ represent the decision-making outputs and the ground-truth ego positions for the future 4 seconds, respectively. $bev_{pr}$ and $bev_{gt}$ are BEV prediction and BEV groundtruth.
$env_{gt}$ is the corresponding static environment map.
The loss weight parameter $\alpha$ is designed to balance the decision loss and the prediction loss.

\subsection{4.4 Online Fine-tuning}

We use Woven Planet L5Kit for autonomous driving simulation to fine-tune the pre-trained model. The BEV inputs are embedded and fed into the model to decide future trajectory points for $M \Delta t$ time intervals. The motion planner is adopted to generate a dynamically feasible trajectory according to the decision outputs. Then the executable trajectory is fed into the model to generate future BEV predictions. Furthermore, an experience rasterizer is developed to help the model with the static information of the simulation driving scenarios. The intuition is that we can easily map all the lanes and intersections into the rasterized BEV image once we know the static global map, the initial world coordinate of the autonomous vehicle, and the transformation between the world coordinate and the raster coordinate.
After $\Delta t$ time interval of simulating, we can obtain the ground-truth of the next BEV image. We fine-tune the model for 10k steps according to Eq. \ref{eq:prediction_loss2}:

\begin{equation}
    \label{eq:prediction_loss2} loss = MSE(env_{er}+bev_{pr}, bev_{sm}), 
\end{equation}
where $env_{er}$, $bev_{pr}$ and $bev_{sm}$ represent the environment map provided by the experience rasterizer, the predicted and simulated BEV image on the L5Kit simulator, respectively.

\section{5 Evaluation}

In this section, we evaluate our framework from a decision-making baseline, motion planning baseline, and driving scenario prediction task.

\subsection{5.1 Decision Making and Motion Planning Metric}
    \label{section:test1}

We evaluate the decision-making ability of our model from the following metrics:
\begin{itemize}
  \item [1)] 
  the final displacement error metric (FDE), which refers to the distance between the predicted position and the reference position at the timestep $t$.
  \item [2)]
  the average displacement error metric (ADE), which refers to the MSE over all predicted positions and the reference positions before the timestep $t$.
  \item [3)]
  the final distance to reference trajectory metric (FDR), which refers to the distance between the predicted position and the closest waypoint in the reference trajectory at the timestep $t$.
  \item [4)]
  the average distance to reference trajectory metric (ADR), which refers to the MSE over all predicted positions and their closest positions before the timestep $t$.
\end{itemize}

We evaluate all the metrics for different prediction horizons.
We compare the performance with the ResNet-based \cite{he2016deep} approach used in \cite{houston2021one}. Only the results of the FDE metric are provided in their paper. Due to the large scene amounts in the validation dataset, we randomly picked 50 scenes from them to test the decision-making baseline. We calculate the average for these metrics.

We evaluate the motion planning capability of our model from the following metrics:
\begin{itemize}
  \item [1)] 
  the L2 error (L2), which refers to the MSE over all positions from the executed trajectories during the simulation and the ground-truth positions from the log records. The simulation duration for each scene is $t$.
  \item [2)]
  the collision rate (CR), which refers to the ratio of the collision scenes within the simulation duration $t$.
  \item [3)]
  the off-road rate (OR), which refers to the ratio of the number of off-road scenes to the total number of scenes. The off-road metric is defined as the distance between the reference trajectory and the autonomous vehicle is greater than 2 meters within the simulation duration $t$.
\end{itemize}

Woven Planet L5Kit is adopted to provide driving simulations for different time intervals. Similarly, we simulate 50 scenes for each interval. 
The performance of our model is compared with another comprehensive framework UniAD ~\cite{hu2023planning} in the L2 metric and the CR metric.

\begin{table*}[ht]
\renewcommand\arraystretch{1.5}
\caption{Decision-Making and Motion Planning Performance Evaluation}
\begin{center}
\begin{small}
\begin{tabular}{ccccccccc}\hline

\textbf{Metrics} & \textbf{Model} & \textbf{@0.5s} & \textbf{@1s} & \textbf{@2s} & \textbf{@3s} & \textbf{@4s} & \textbf{Average} \\\hline
ADE & BEVGPT(ours)  & \textbf{0.17} & \textbf{0.31} & \textbf{0.59} & \textbf{0.86} & \textbf{1.16} & \textbf{0.62} \\ 
FDE & BEVGPT(ours)  & \textbf{0.27} & \textbf{0.56} & \textbf{1.13} & \textbf{1.56} & \textbf{2.17} & \textbf{1.14} \\ 
FDE & ResNet-based \cite{houston2021one} 
& 0.31 & 0.59 & 1.10 & 1.61 & 2.14 & 1.15  \\ 
FDR & BEVGPT(ours)  & \textbf{0.22} & \textbf{0.34} & \textbf{0.51} & \textbf{0.64} & \textbf{0.97} & \textbf{0.54} \\ 
ADR & BEVGPT(ours)  & \textbf{0.14} & \textbf{0.22} & \textbf{0.34} & \textbf{0.42} & \textbf{0.52} & \textbf{0.33} \\ 
L2 & BEVGPT(ours)  & \textbf{0.24} & \textbf{0.39} & \textbf{0.88} & \textbf{1.70} & \textbf{2.87} & \textbf{1.22} \\ 
L2 & UniAD \cite{hu2023planning}  & - & 0.48 & 0.96 & 1.65 & - & 1.03 \\ 
CR & BEVGPT(ours)  & \textbf{0.00} & \textbf{0.00} & \textbf{0.16} & \textbf{0.20} & \textbf{0.40} & \textbf{0.15} \\ 
CR & UniAD \cite{hu2023planning}  & - & 0.05 & 0.17 & 0.71 & - & 0.31 \\ 
OR & BEVGPT(ours)  & \textbf{0.00} & \textbf{0.00} & \textbf{0.06} & \textbf{0.20} & \textbf{0.26} & \textbf{0.10} \\ 
\hline
\end{tabular}
\end{small}
\label{tbl:decision_evaluation}
\end{center}
\end{table*} 

The evaluation results are summarized in Table \ref{tbl:decision_evaluation}. The results of our model are highlighted with bold fonts. Our model outperforms the ResNet-based method from \cite{houston2021one} in the final displacement error metric (FDE) and also demonstrates excellent performance in other decision-making metrics. Compared with UniAD \cite{hu2023planning}, our model presents surprising capability in the off-road rate metric and the collision rate metric, since we do not explicitly add the driving safety and obstacle avoidance constraints in the motion planning module.

\subsection{5.2 Driving Scenario Prediction}
    \label{section:test2}

To evaluate the capability of long future driving scenario prediction, we design the experiments where the model is required to generate future BEV images within $T=6s$. 
In this case, BEVGPT is required to auto-regressively predict future driving scenarios over 60 timesteps, with the first frame's BEV ground-truth.
We choose four challenging traffic scenarios. The first one is an intersection with a traffic light, where the vehicle should understand the meaning of the red light and stop before the intersection. The second is a highly-dynamic intersection with a green light, where the vehicle needs to pass the intersection correctly.
The third is a straight road in a multi-agent environment, where the autonomous vehicle should drive forward as fast as possible.  
The last one is the road intersection, where the traffic signal changes from red to green, and the vehicle needs to comprehend this information and start promptly.

\begin{figure*}[!ht]
  \centering
  \includegraphics[width=0.95\textwidth]{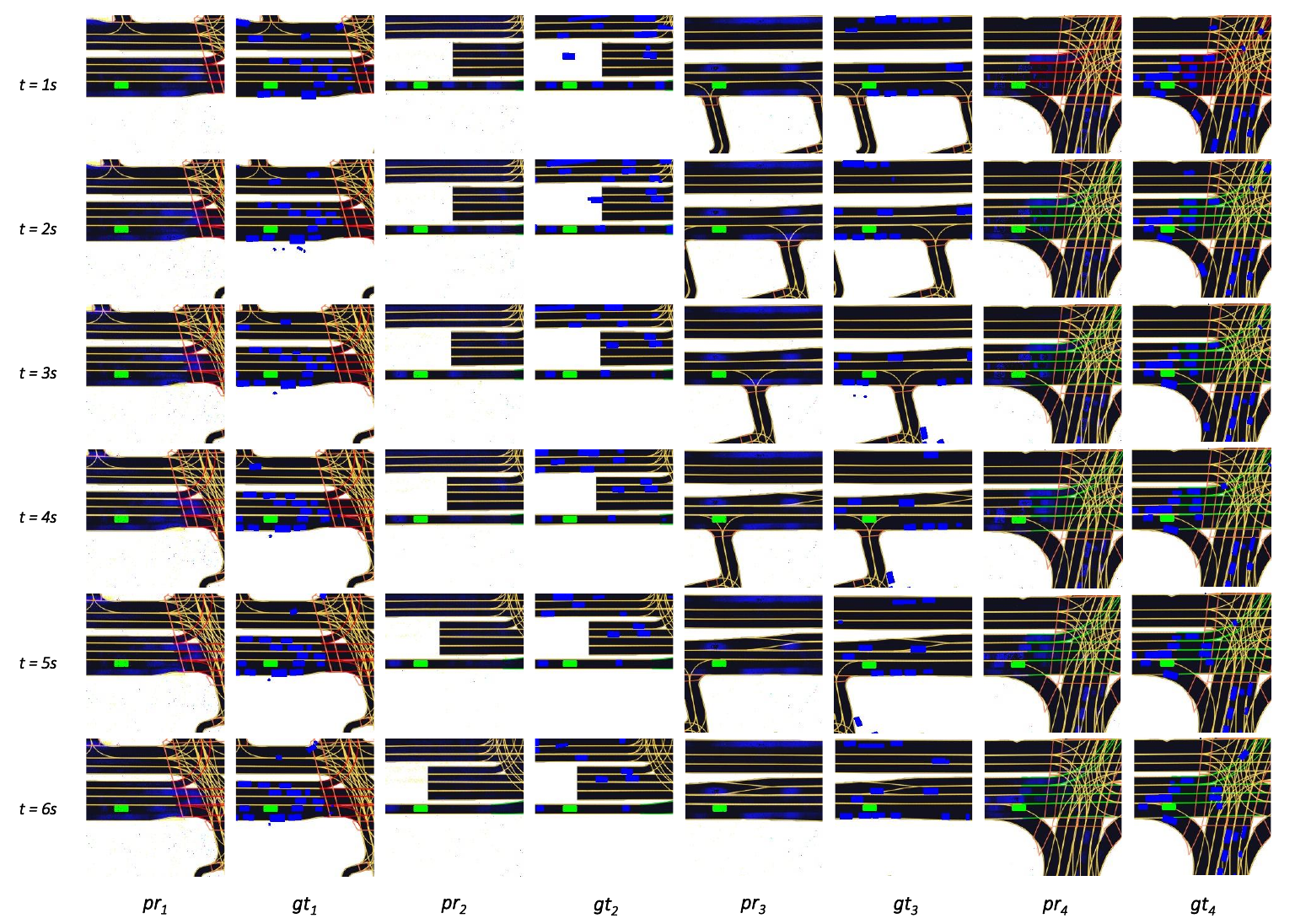}
  \caption{The driving scenario prediction and ground truth for each scene over a duration $T=6s$. $pr_1$ and $gt_1$ refer to the prediction and ground truth for the first scene, while $pr_2$ and $gt_2$ refer to the prediction and ground truth for the second scene, respectively.}\label{fig:bev_prediction}
\end{figure*}

Figure \ref{fig:bev_prediction} shows the driving scenario prediction and ground truth for each scene over a duration $T=6s$. 
The ability of our framework to predict long-term future driving scenarios is demonstrated by accurate BEV generation.

\section{6 Discussion and Conclusion}

In this paper, we consider the minimalist composition of an autonomous driving system from the perspective of human drivers and apply the bird's-eye-view (BEV) representation as the unified input.
Therefore, we propose BEVGPT, a comprehensive GPT model that integrates driving scenario prediction, decision-making, and motion planning. We employ a two-stage training procedure. We first use a large amount of self-driving data from the Lyft Level 5 Dataset \cite{houston2021one} to pre-train the causal transformer. Subsequently, we adapt the model by online learning using an online autonomous driving simulator L5Kit. We develop a motion planner that takes control effort into consideration while following the state constraints, the continuity constraints, and the dynamic constraints. Moreover, an experience rasterizer is designed to help the model with the static information of the simulation driving scenarios.

We design a variety of metrics, pick large amounts of driving scenarios and execute numerous realistic simulations to evaluate the decision-making and motion planning performance of BEVGPT. The evaluation results demonstrate that our model exceeds previous methods in most metrics while performing similarly in other metrics.
Furthermore, we wish to evaluate how well our model works under certain extreme conditions, such as auto-regressively predicting the future driving scenarios over 60 frames according to the first frame's BEV image ground-truth as the model input. 
Surprisingly, our model performs well with such difficult challenges.
As a result, the ability of our framework to predict long-term future driving scenarios is demonstrated by accurate BEV generation.

To sum up, our study incorporates the decision-making, motion planning, and driving scenario prediction modules into a minimalist autonomous driving framework. It demonstrates the practicability of the BEV representation as the unified input. Moreover, BEVGPT provides a new paradigm for autonomous driving where all these tasks are incorporated into a single GPT model with only BEV images as input. An additional meaning of our study is the successful deployment of the GPT architecture in autonomous driving systems. The outstanding performance reveals the enormous potential of GPT structures in such complex and highly dynamic systems.

\section{Acknowledgments}
This study is supported by the National Natural Science Foundation of China under Grant 52302379, Guangzhou Basic and Applied Basic Research Project 2023A03J0106, Guangdong Province General Universities Youth Innovative Talents Project under Grant 2023KQNCX100, and Guangzhou Municipal Science and Technology Project 2023A03J0011.

\bibliography{aaai24}

\end{document}